\documentclass[letterpaper, 10 pt, conference]{ieeeconf}  

\IEEEoverridecommandlockouts                              

\usepackage{amsmath}
\usepackage{amssymb}   
\usepackage{graphicx}
\usepackage{caption}
\usepackage{subcaption}
\usepackage{float}
\usepackage{booktabs} 
\usepackage{multirow}
\usepackage{makecell}
\usepackage{array}
\usepackage{algorithm}
\usepackage{algpseudocode} 
\usepackage{amsmath} 
\usepackage{amsmath} 
\usepackage{algorithm}
\usepackage{algpseudocode}

\usepackage{graphicx}
\usepackage{booktabs}
\usepackage{multirow}
\usepackage{adjustbox} 
\usepackage{colortbl}  
\usepackage{xcolor}
\definecolor{tabblue}{rgb}{0.88, 0.92, 0.98} 

\title{\LARGE \bf
GeoLAM: Learning Geometry-Grounded Latent Actions from Unlabeled Human Videos
}

\author{Yifan Xie$^{1*}$, Hekun Tian$^{1*}$, Jinkun Liu$^{1}$,
YuAn Wang$^{2}$, Qiao Sun$^{3}$, and Wenbo Ding$^{1\dagger}$%
\thanks{$^{*}$These authors contributed equally to this work.}%
\thanks{$^{1}$Tsinghua University.}%
\thanks{$^{2}$Beijing Institute of Technology.}%
\thanks{$^{3}$Fudan University.}%
\thanks{$^{\dagger}$Corresponding author: Wenbo Ding.}%
}

\begin{document}

\IEEEaftertitletext{\vspace{-\baselineskip}}
\makeatletter
\let\GeoLAMoriginalmakefntext\@makefntext
\long\def\@makefntext#1{\parindent\z@\noindent\hbox{\@makefnmark}#1}
\maketitle
\let\@makefntext\GeoLAMoriginalmakefntext
\makeatother
\thispagestyle{empty}
\pagestyle{empty}
\raggedbottom 

\begin{abstract}
Human videos provide rich manipulation experience, but extracting action
representations that preserve useful motion remains challenging. Visual
reconstruction alone can entangle manipulation-related motion with
appearance changes and camera movement. We present GeoLAM, a framework for
learning geometry-grounded latent actions from action-free human videos.
GeoLAM combines future-frame reconstruction through a frozen geometric
feature hierarchy with motion supervision from a training-only 4D geometry
teacher. The geometric representation provides a structural prior, while
the teacher's predictions yield spatially pooled targets capturing 3D
displacement, residual image-plane motion, and surface-orientation changes.
Visibility and confidence weighting reduces the contribution of unreliable
estimates, encouraging continuous latent actions to retain geometric motion
without explicit hand-pose or hand-trajectory annotations. After video
pretraining without action labels, the learned representation provides
transition targets for a world-action model trained on action-labeled robot
demonstrations. The model jointly denoises latent actions and executable
action chunks, with future-video prediction used only as an auxiliary
training task. Deployment therefore requires neither the geometry teacher
nor future-video generation. Evaluations on a latent-action benchmark and
robotic manipulation tasks demonstrate the strong performance of GeoLAM.

\end{abstract}


\section{INTRODUCTION}

Imitation learning has enabled fine-grained robot manipulation using
demonstrations paired with executable actions~\cite{zhao2023act}.
Human videos offer a complementary source of everyday interaction experience
at scale~\cite{grauman2022ego4d}, but typically lack corresponding robot
commands. Explicit motion reconstruction provides one route to recovering
supervision: methods such as HaWoR~\cite{zhang2025hawor} estimate world-space hand poses and
trajectories from egocentric videos. However, occlusion and tracking failures
complicate reliable motion estimation, and reconstructing hand and camera
motion incurs additional preprocessing computation.
The resulting human-motion estimates also require a mapping to the target
robot's control space. These challenges motivate learning an intermediate
action representation directly from visual transitions, without requiring
explicit hand-pose or hand-trajectory annotations. Latent-action models offer
this possibility by inferring compact transition variables from video and
subsequently connecting them to executable
actions~\cite{ye2025lapa}.

The usefulness of such a representation depends on which aspects of a visual
transition it preserves. Inverse and forward dynamics models can jointly
learn latent actions through future-observation prediction, yet reconstruction
alone does not ensure that the latent variables retain motion relevant to
manipulation. Appearance variation, camera movement, and object motion can
all contribute to the prediction objective, leaving the content of the
action bottleneck ambiguous.
Recent optical-flow-constrained and 3D-aware latent-action methods highlight
the value of explicitly incorporating motion and
geometry~\cite{bu2026laof,yang2026lawm3d}. Building on this direction, we focus
on two complementary requirements: a geometric prior for the visual state
and a direct constraint on the motion encoded by the latent transition.
Geometric foundation models provide representations informed by scene
structure~\cite{lin2026depthanything3}, while a temporal learning signal can
encourage the transition variable to preserve changes in that structure.
For manipulation, we seek compact latent actions that retain spatial
displacement and surface-orientation changes while reducing the influence of
unreliable observations and static background.

We introduce \textbf{GeoLAM}, a framework for learning geometry-grounded
latent actions from action-free human videos and using them for robot
control. GeoLAM combines future-frame reconstruction through a frozen
geometric feature hierarchy with explicit geometric motion supervision.
The first constraint places transition modeling in a geometry-informed
representation space. The second uses a training-only 4D
teacher~\cite{zhang2026d4rt} to construct motion targets from predicted
3D displacements, residual image-plane motion, and surface-orientation
changes. Visibility- and confidence-weighted spatial pooling reduces the
contribution of unreliable estimates, providing geometric guidance without
requiring explicit human-hand reconstruction. We pretrain this continuous
latent-action representation on approximately 20{,}000 hours of human video
without using action annotations. During subsequent training on
action-labeled robot demonstrations, a world-action model jointly denoises
latent actions and robot action chunks~\cite{li2026lawa}, using GeoLAM's
representation as the transition target. Future-video prediction remains
an auxiliary training task~\cite{yuan2026fastwam}. Deployment requires
neither the geometry teacher nor future-video generation. Evaluations on
LARYBench~\cite{nie2026lary} and robotic manipulation tasks demonstrate
the strong performance of GeoLAM.

\section{RELATED WORK}
\subsection{Latent Action Model for Manipulation}
Latent-action models recover compact transition variables from videos by
coupling inverse and forward dynamics, providing an intermediate
representation for subsequent policy learning~\cite{ye2025lapa}.
Existing studies differ in whether robot videos are included and whether
robot actions supervise the latent space. Using robot videos without their
action labels remains action-free, making this setting distinct from
pretraining on human videos alone. LAPA~\cite{ye2025lapa} learns
quantized latent actions without action labels and evaluates both robot-video
and human-only pretraining. Other approaches mix human and robot videos,
adding auxiliary prediction of robot states and actions when annotations
are available~\cite{chen2025villa}, or supervise latent dynamics with
end-effector motion regression on action-labeled frames~\cite{shen2026ld4wam}.
Broader robot-learning pipelines also exploit action-labeled manipulation
trajectories~\cite{team2026xiaomi} and reconstructed human-motion
priors~\cite{xie2026humanintention}. World-action models further connect
prediction to control through structured object states with persistent
identities~\cite{liu2026oawam} or latent transition variables jointly
denoised with robot action chunks~\cite{li2026lawa}. Future-video prediction
can also serve as an auxiliary training objective, allowing deployment
without explicit future-video generation~\cite{yuan2026fastwam}.

GeoLAM learns its latent-action encoder exclusively from action-free human
RGB videos, without robot videos, robot action labels, or explicit
hand-motion annotations. Reconstruction and teacher-derived geometric
motion provide the learning signals. The encoder remains frozen during
downstream training on action-labeled robot demonstrations, so policy
learning uses its latent targets without reshaping the pretrained
representation.

\subsection{Geometric Foundation Model for Manipulation}
Geometric and predictive representations provide structural priors for
understanding physical scenes. Predicting target-block embeddings from
point-cloud context with a context-aware decoder supports transferable
3D representations~\cite{hu2024jepa3d}, while geometric foundation models
recover scene structure across RGB views~\cite{lin2026depthanything3}.
Such pretrained features have been
repurposed for multi-view generation~\cite{jang2026gld} and robot policies
that integrate perception, temporal prediction, and action
decoding~\cite{han2026gam}. Broader studies of embodied interaction also investigate
reasoning about the physical properties of objects~\cite{xie2025universal}
and local action correction using interaction
feedback~\cite{li2026mastermicro}. For latent-action learning,
optical-flow supervision encourages the bottleneck to encode observable
motion~\cite{bu2026laof}, while geometric feature alignment, RGB-D future
reconstruction, and learning across viewpoints introduce 3D structural
constraints~\cite{yang2026lawm3d}. Dynamic reconstruction models extend
geometric prediction across time, providing point correspondences and
surface normals together with visibility and confidence
estimates~\cite{zhang2026d4rt}. These capabilities offer both spatial
representations and temporal geometric cues for modeling interactions.

GeoLAM combines reconstruction through a frozen geometric feature hierarchy
with explicit supervision of 3D displacement, residual image-plane motion,
and surface-orientation changes. Visibility- and confidence-weighted
teacher targets constrain the latent transition during human-video
pretraining. The geometry teacher is unnecessary for downstream policy
training or deployment.


\begin{figure*}[!t]
    \centering
    \includegraphics[width=0.9\textwidth]{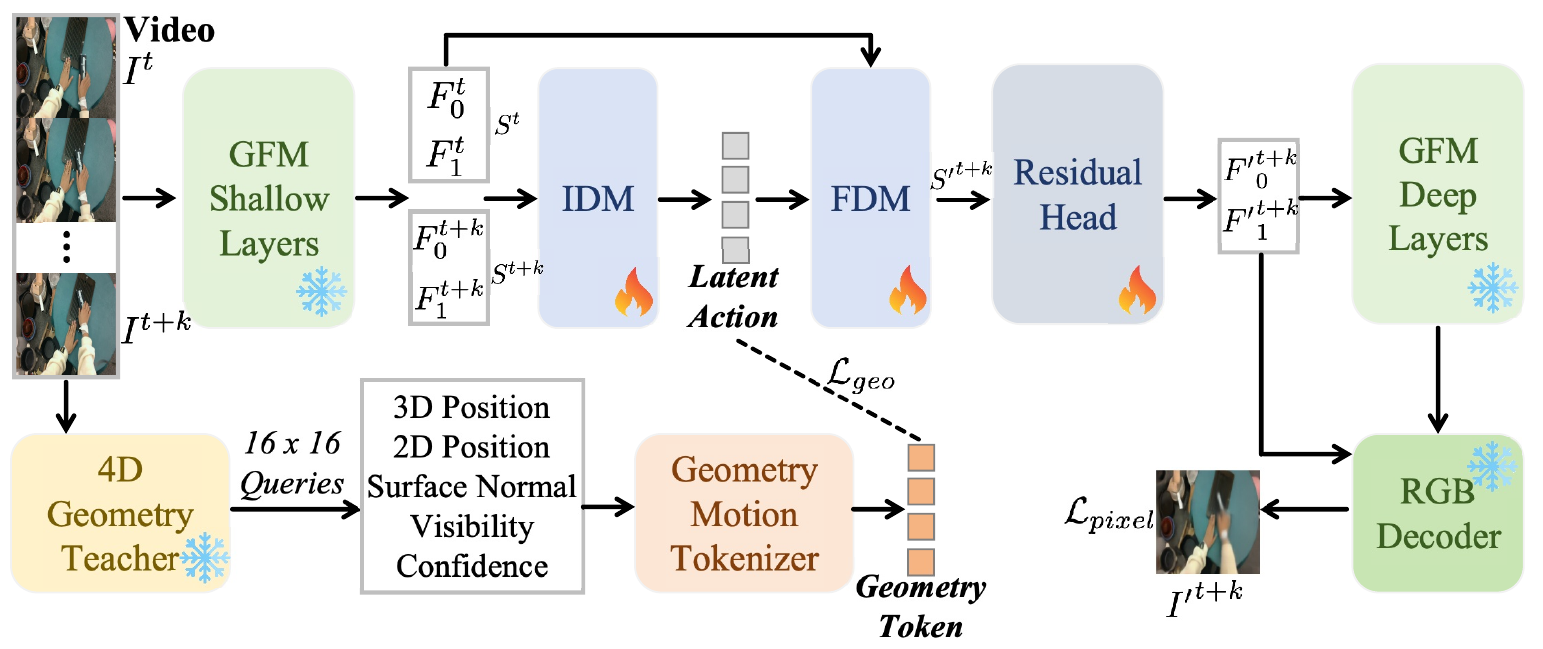}
    \caption{\textbf{Overview of GeoLAM.}
    Frozen shallow GFM features encode the two video endpoints. The inverse dynamics model (IDM) compresses their transition into latent-action tokens, and the forward dynamics model (FDM) predicts future features from the current state and these tokens. Frozen deep GFM layers and an RGB decoder provide the reconstruction objective $\mathcal L_{\mathrm{pixel}}$. During training, a frozen 4D teacher converts reliable 3D/2D motion and surface-normal changes into geometry-motion targets for $\mathcal L_{\mathrm{geo}}$. Snowflakes and flames
    denote frozen and trainable modules, respectively.}
    \label{fig:lam}
\end{figure*}

\section{Method}\label{sec:method}

\subsection{Geometric Latent Action Learning}
\label{sec:geometric_lal}
Latent-action models recover transition variables from action-free videos by coupling an inverse dynamics model (IDM) with a forward dynamics model (FDM)~\cite{ye2025lapa}. However, image prediction alone may entangle controllable motion with appearance changes and camera motion~\cite{bu2026laof,yang2026lawm3d}. As illustrated in Fig.~\ref{fig:lam}, we reduce this ambiguity with two complementary constraints: geometry-aware future-frame reconstruction and training-only supervision from a 4D geometry teacher. We denote the two clip endpoints by $\mathbf I^t$ and $\mathbf I^{t+k}$.

\subsubsection{3D Latent Representation}
Rather than modeling dynamics directly in RGB space, we construct the visual state in the feature hierarchy of a frozen geometric foundation model (GFM). We instantiate the hierarchy with the Depth Anything 3 backbone and the multi-level reconstruction decoder of GLD~\cite{lin2026depthanything3,jang2026gld}, following the broader principle of repurposing geometry-pretrained representations for action-conditioned prediction~\cite{han2026gam}. For each endpoint $\tau$, the frozen shallow GFM yields $\mathbf F_{0:1}^{\tau}=\operatorname{sg}(\mathcal E_{\mathrm{gfm}}^{0:1}(\mathbf I^\tau))$, which a learned state encoder maps to $\mathbf S^\tau=\mathcal E_s(\mathbf F_{0:1}^{\tau})$. The IDM represents the transition as $\mathbf M^{t\rightarrow t+k}=\mathcal T_{\mathrm{idm}}(\phi_m(\mathbf S^{t+k}-\mathbf S^t))$. Position-aware action queries then aggregate these transition features, and a deterministic bottleneck produces
\begin{equation}
    \mathbf Z^t
    =\mathcal B_{\mathrm{det}}\!\left(
      \operatorname{CA}(\mathbf Q_a+\mathbf P_a,
      \mathbf M^{t\rightarrow t+k})\right).
    \label{eq:latent_representation}
\end{equation}
Here $\operatorname{sg}$ denotes stop-gradient and $\mathcal B_{\mathrm{det}}$ compresses and lifts each action token without sampling or vector quantization. Operating on the state difference, rather than concatenating the endpoints, focuses the bottleneck on transition information.

The FDM conditions the current state on the complete latent-action token set to obtain $\widehat{\mathbf S}^{t+k}=\mathcal T_{\mathrm{fdm}}(\mathbf S^t;\mathbf Z^t)$. Each residual head updates a shallow feature as $\widehat{\mathbf F}_\ell^{t+k}=\mathbf F_\ell^t+h_\ell(\widehat{\mathbf S}^{t+k})$. The frozen deep GFM propagator and RGB decoder then reconstruct $\widehat{\mathbf I}^{t+k}$ from the predicted shallow features, yielding the photometric loss $\mathcal L_{\mathrm{pixel}}=\operatorname{MSE}(\widehat{\mathbf I}^{t+k},\mathbf I^{t+k})$. Both frozen modules remain differentiable with respect to the predicted features. This branch therefore needs neither depth labels nor camera parameters: its geometric inductive bias is inherited from the pretrained GFM.

\subsubsection{4D Geometric Teacher}
Pixel reconstruction can exploit appearance cues without preserving physical scene motion. We therefore use a frozen implementation of D4RT~\cite{zhang2026d4rt} as a training-only, output-space geometry teacher. It processes a uniformly sampled clip between the endpoints, whereas the latent-action model observes only $\mathbf I^t$ and $\mathbf I^{t+k}$. Paired grid queries share the source point and reference camera but target the initial and final times. From their predicted 3D/2D positions, surface normals, visibility, and confidence, we construct interpretable motion targets without exposing hidden D4RT features. Unlike flow-only supervision~\cite{bu2026laof}, these targets also retain 3D displacement and surface-orientation change.

We subtract the component-wise median from endpoint image-plane displacement to obtain residual motion $\mathbf r_i$, and derive a reliability weight $\omega_i$ from endpoint visibility and confidence. The per-query descriptor comprises bounded, scene-normalized 3D displacement $\boldsymbol\delta_i^{3d}$, residual 2D motion $\boldsymbol\delta_i^{2d}$, normalized surface-orientation change $\delta_i^n$, and residual-motion magnitude $\delta_i^m$.

The tokenizer concatenates these cues as $\boldsymbol\phi_i=[\boldsymbol\delta_i^{3d};\boldsymbol\delta_i^{2d};\delta_i^n;\delta_i^m]$ and pools them over coarse spatial regions:
\begin{equation}
    \mathbf m_j^\star
    =\frac{\sum_{i\in\mathcal R_j}a_i\boldsymbol\phi_i}
           {\sum_{i\in\mathcal R_j}a_i}.
\label{eq:geometry_tokens}
\end{equation}
Here $a_i=\omega_i g(\|\mathbf r_i\|_2)$, and the bounded increasing function $g$ emphasizes observable motion. Stacking the regional tokens gives the teacher target $\mathbf M^\star$. This reliability- and motion-weighted pooling suppresses uncertain tracks and static background while retaining coarse spatial organization.

A lightweight position-aware projector cross-attends to all latent-action tokens and predicts $\widehat{\mathbf M}=h_{\mathrm{geo}}(\mathbf Z^t)$. It is supervised by $\mathcal L_{\mathrm{geo}}=\operatorname{MSE}(\widehat{\mathbf M},\operatorname{sg}(\mathbf M^\star))$. The complete training objective is
\begin{equation}
    \mathcal L(n)
    =\lambda_{\mathrm{pixel}}\mathcal L_{\mathrm{pixel}}
     +\lambda_{\mathrm{geo}}(n)\mathcal L_{\mathrm{geo}}.
\label{eq:training_objective}
\end{equation}
We first optimize reconstruction alone and then linearly increase the geometry weight. The teacher is never provided to the FDM and is discarded at inference. It affects the latent action only through $\mathcal L_{\mathrm{geo}}$, encouraging geometry-predictive motion without adding test-time cost.

\subsection{Latent Action for World Action Model}
\label{sec:latent_wam}
World-action models can generate future observations before control
~\cite{shen2026ld4wam} or use future prediction only during training
~\cite{yuan2026fastwam}. We follow the efficient deployment pattern of
Fast-WAM while giving the action generator an explicit transition variable:
the continuous latent action from Sec.~\ref{sec:geometric_lal} is jointly
denoised with the robot action chunk, and future-video prediction remains an
auxiliary training task. Unlike codebook-derived targets
~\cite{ye2025lapa,li2026lawa} or dynamics partially aligned with labeled
end-effector motion~\cite{shen2026ld4wam}, our target is a deterministic,
geometry-grounded transition summary learned from action-free video.

\subsubsection{Transition Targets and Visual Context}
\begin{figure}[!t]
    \centering
    \includegraphics[width=1.0\linewidth]{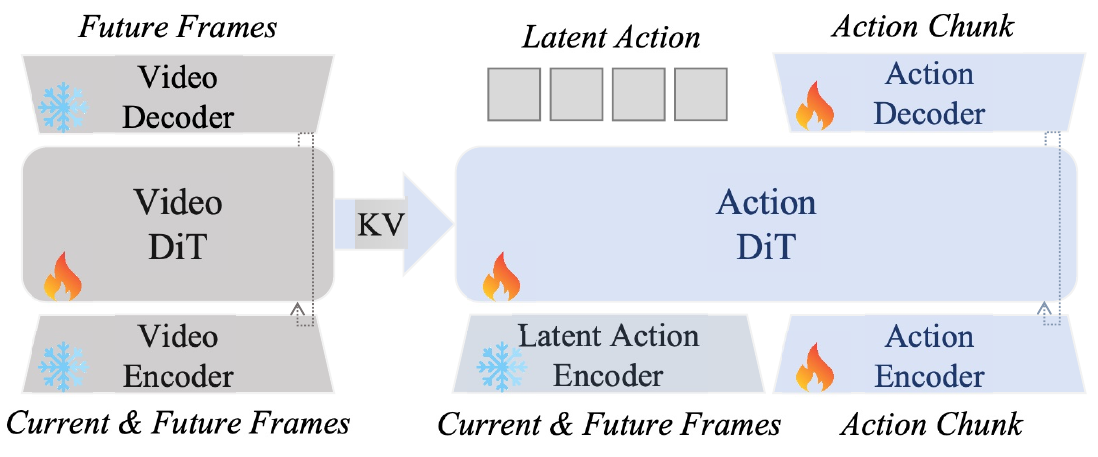}
    \caption{\textbf{Geometry-grounded latent actions for the world action
    model:} during training, a frozen encoder supplies continuous transition
    targets and future-video prediction regularizes the Video DiT, whereas
    inference jointly denoises latent actions and action chunks using only
    cached current-observation key--value features. Snowflakes and flames
    denote frozen and trainable modules, respectively.}
    \label{fig:wam}
\end{figure}

Figure~\ref{fig:wam} shows the two-branch training architecture. For a robot
demonstration, let $\mathbf A^t=[\mathbf a^t,\ldots,\mathbf a^{t+H-1}]$ be an
$H$-step action chunk and let
$\mathbf I_+^t=[\mathbf I^{t_1},\ldots,\mathbf I^{t_T}]$, with
$t<t_1<\cdots<t_T=t+k$, contain its future frames. The terminal frame
$\mathbf I^{t+k}$ is aligned with the state reached after the chunk. The values of $k$ and
$H$ may differ because camera and control rates need not match. The four
spatial latent-action queries summarize this interval and are not paired
one-to-one with video frames or action steps. We denote the task embedding by
$\boldsymbol\ell^t$ and the current proprioceptive state by $\mathbf q^t$.

During policy training, we freeze the latent-action encoder
$\mathcal E_{\mathrm{la}}$, comprising the GFM shallow layers and the trained
IDM from Sec.~\ref{sec:geometric_lal}. It provides the target
$\mathbf Z_\star^t=\operatorname{sg}(\mathcal E_{\mathrm{la}}(
\mathbf I^t,\mathbf I^{t+k}))$. A frozen framewise encoder $\mathcal E_v$
separately produces the current latent
$\mathbf V_0^t=\operatorname{sg}(\mathcal E_v(\mathbf I^t))$ and future
latents $\mathbf V_+^t=\operatorname{sg}(\mathcal E_v(\mathbf I_+^t))$,
preventing future information from entering the current anchor before
attention. The FDM and geometry teacher supervise the IDM only in the
preceding stage and are not loaded during world-action model training.

Following Fast-WAM~\cite{yuan2026fastwam}, the future branch is trained but
omitted at test time. A DiT-style video denoiser~\cite{peebles2023dit} applies
conditional flow matching~\cite{lipman2023flow} to $\mathbf V_+^t$ while
keeping $\mathbf V_0^t$ clean. For $s\sim\mathcal U[0,1]$, it receives
$\mathbf V_{+,s}^t=(1-s)\mathbf V_+^t+s\boldsymbol\epsilon_v$ and predicts the
target velocity
$\mathbf u_{\mathrm{vid}}^t=\boldsymbol\epsilon_v-\mathbf V_+^t$ as
$\widehat{\mathbf u}_{\mathrm{vid}}^t$. A
block-causal mask lets future tokens read the anchor but prevents the anchor
from reading the future. The anchor uses a fixed clean-time embedding, and its
layer-wise keys and values form $\mathbf C^t$, the only Video-DiT features
exposed to the Action DiT. Hence $\mathbf C^t$ depends on the current
observation and $\boldsymbol\ell^t$, but not on the ground-truth future or
$s$. The frozen video decoder is used only for qualitative visualization.

\subsubsection{Coupled Transition--Control Flow}
The Action DiT operates on two jointly denoised token sets. We adapt the joint
latent--action denoising principle of LAWA~\cite{li2026lawa}, but replace its
codebook-derived target with our continuous geometric representation. After
standardizing both modalities with training-set statistics, we use one shared
$\tau\sim\mathcal U[0,1]$. For
$m\in\{\mathrm{lat},\mathrm{act}\}$, let
$\mathbf x^{\mathrm{lat}}=\mathbf Z_\star^t$ and
$\mathbf x^{\mathrm{act}}=\mathbf A^t$. The noisy input is
$\mathbf x_\tau^m=(1-\tau)\mathbf x^m+\tau\boldsymbol\epsilon_m$ and its
target velocity is $\mathbf u_m=\boldsymbol\epsilon_m-\mathbf x^m$, with
independent Gaussian noise for the two modalities.

A learned projection embeds the noisy latent tokens, while the action encoder
embeds the noisy action chunk. Modality-specific positional embeddings
distinguish the spatial transition queries from the temporal action tokens.
The Action DiT conditions both sets on $\tau$, $\mathbf C^t$,
$\boldsymbol\ell^t$, and $\mathbf q^t$. Latent tokens read only visual and
latent tokens, whereas action tokens can additionally read the action stream.
This prevents direct access to demonstrated actions when predicting the latent
transition. Separate heads predict the latent and action velocities, with the
action decoder serving as the velocity head at each integration step.

For each $m\in\{\mathrm{vid},\mathrm{lat},\mathrm{act}\}$, conditional flow
matching minimizes
$\mathcal L_m=\mathbb E[\|\widehat{\mathbf u}_m-\mathbf u_m\|_2^2]$ over valid
tokens. The complete objective is
\begin{equation}
    \mathcal L_{\mathrm{wam}}
      =\lambda_{\mathrm{vid}}\mathcal L_{\mathrm{vid}}
       +\lambda_{\mathrm{lat}}\mathcal L_{\mathrm{lat}}
       +\lambda_{\mathrm{act}}\mathcal L_{\mathrm{act}}.
\label{eq:wam_objective}
\end{equation}
Padded steps are masked. The latent term aligns the predicted transition with
the geometry-aware target, while the action term grounds it in executable
control. All three losses update the Video DiT, with the latter two propagating
through $\mathbf C^t$. Only the latent and action losses update the Action DiT.
Both encoders and the video decoder remain frozen.

At inference, only the current observation is encoded. The Video DiT performs
one fixed-clean-time anchor pass to populate $\mathbf C^t$. The future-video
stream, frozen video decoder, and latent-action target encoder are absent.
Starting from Gaussian latent-action and action variables, we reuse this cache
while integrating the coupled flow from $\tau=1$ to $\tau=0$. The terminal
action state is inverse-standardized to obtain the action chunk. The model thus
predicts multi-step controls as in action-chunking policies~\cite{zhao2023act}.
Following receding-horizon execution, only a short
prefix is executed before replanning. The predicted latent action is therefore
an online transition estimate for control rather than a robot command or a
generated future frame.

\section{EXPERIMENTS}
\label{sec:exp}

\subsection{Implementation Details}

\noindent\textbf{Training.}
All models are trained with distributed data parallelism on 64 NVIDIA A800 GPUs using bfloat16 mixed precision. We optimize the latent-action model for 200{,}000 steps with AdamW, a learning rate of $10^{-4}$, and a per-GPU batch size of 32 (2{,}048 globally). RGB inputs are resized to $224\!\times\!224$, while the frozen D4RT teacher processes four uniformly sampled context frames at $256\!\times\!256$. The IDM and FDM each use eight Transformer blocks. The IDM outputs four latent-action tokens through a 32-dimensional deterministic bottleneck. We train with $\mathcal L_{\mathrm{pixel}}$ alone for 10{,}000 steps, then linearly ramp the geometry-loss weight from 0 to 1 over 1{,}000 steps. For WAM training, we initialize the video branch from Wan2.2-5B~\cite{wan2025} and use nine-frame clips, an action expert with hidden width 1{,}024, and horizon $H=32$. Following Fast-WAM~\cite{yuan2026fastwam}, future-video prediction remains training-only, while the latent action is jointly denoised with the executable action chunk.

\begin{table*}[!t]
    \centering
    \caption{Results on the control regression and composite action classification tasks of LARYBench~\cite{nie2026lary}.
    Human and Robot denote the composite classification subsets. Accuracy is in percent.
    Best results among the compared methods are in bold.}
    \label{tab:larybench}
    \small
    \setlength{\tabcolsep}{4pt}
    \renewcommand{\arraystretch}{1.10}
    \begin{tabular*}{\textwidth}{@{\extracolsep{\fill}}l*{5}{c}@{\hspace{10pt}}*{3}{c}@{}}
        \toprule
        \multirow{2}{*}{\textbf{Model}}
        & \multicolumn{5}{c}{Regression MSE $\downarrow$}
        & \multicolumn{3}{c}{Classification Accuracy $\uparrow$} \\
        \cmidrule(lr){2-6} \cmidrule(lr){7-9}
        & CALVIN & VLABench & RoboCOIN & AgiBot & Avg.
        & Human & Robot & Avg. \\
        \midrule
        LAPA~\cite{ye2025lapa} & 0.96 & 0.95 & 0.96 & 1.00 & 0.97 & 14.61 & 23.64 & 19.13 \\
        UniVLA~\cite{bu2025univla} & 0.82 & 0.74 & 0.94 & 0.97 & 0.87 & 19.08 & 18.56 & 18.82 \\
        villa-X~\cite{chen2025villa} & 0.86 & 0.72 & 0.94 & 0.97 & 0.87 & 17.80 & 29.90 & 23.85 \\
        LAPA-DINOv3~\cite{nie2026lary} & 0.50 & 0.25 & 0.82 & 0.84 & 0.60 & 64.19 & 27.04 & 45.62 \\
        \midrule
        \textbf{Ours} & \textbf{0.27} & \textbf{0.09} & \textbf{0.40} & \textbf{0.42}
        & \textbf{0.29} & \textbf{74.19} & \textbf{68.25} & \textbf{71.22} \\
        \bottomrule
    \end{tabular*}
\end{table*}

\begin{figure}[!t]
    \centering
    \includegraphics[width=0.8\linewidth]{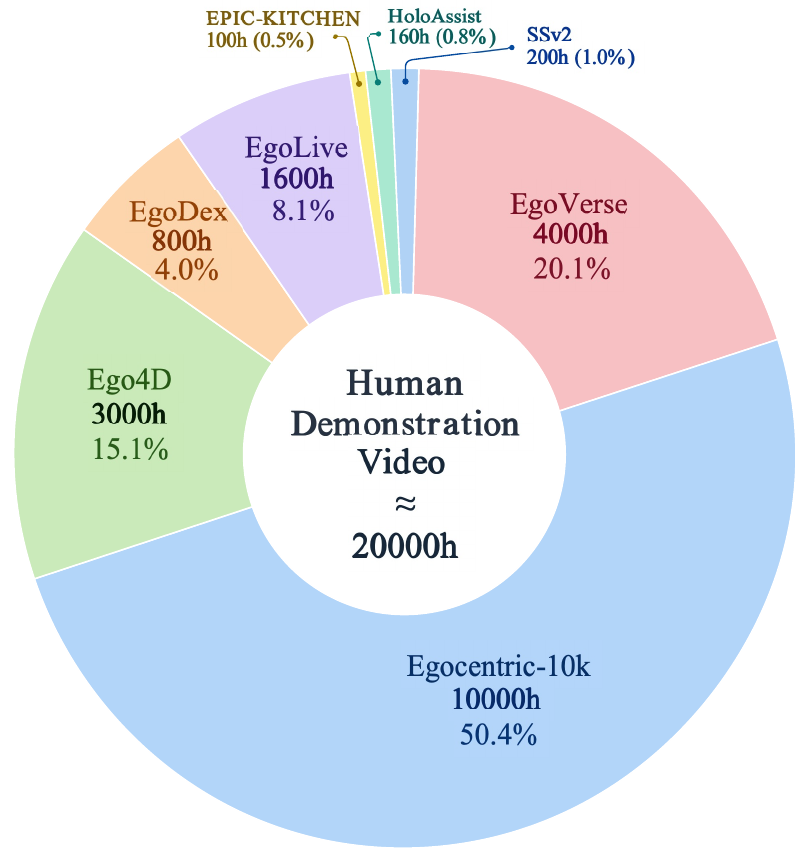}
    \caption{\textbf{Composition of the human-video pretraining corpus.}
    The selected clips total 19{,}860 hours (approximately 20{,}000 hours). Labels report the rounded duration and mixture share from each source.}
    \label{fig:data}
\end{figure}

\noindent\textbf{Data.}
Our action-free pretraining corpus contains 19{,}860 hours of human video selected from Egocentric-10K~\cite{buildai2025egocentric10k}, EgoVerse~\cite{punamiya2026egoverse}, Ego4D~\cite{grauman2022ego4d}, and EgoLive~\cite{li2026egolive}, together with EgoDex~\cite{hoque2026egodex}, Something-Something V2~\cite{goyal2017something}, HoloAssist~\cite{wang2023holoassist}, and EPIC-KITCHENS-100~\cite{damen2022epickitchens}. As summarized in Fig.~\ref{fig:data}, Egocentric-10K contributes roughly half of the mixture, while the remaining sources broaden its coverage of everyday, dexterous, and task-oriented interactions. We use RGB only, sample endpoint pairs at temporal strides in $\{5,10,15,20\}$, and consume neither annotations nor camera parameters.

\subsection{Latent Action Benchmark}
\label{sec:latent_action_benchmark}

To assess whether our latent actions capture both fine-grained motion and
high-level action semantics, we evaluate GeoLAM on LARYBench~\cite{nie2026lary}.
Its probe-based evaluation measures how effectively latent representations
support low-level control regression and semantic action classification.
We report regression MSE on CALVIN, VLABench, RoboCOIN, and AgiBot, together
with composite action classification accuracy on the human and robot subsets.

As shown in Table~\ref{tab:larybench}, GeoLAM achieves the lowest regression
MSE on all four datasets and the highest classification accuracy on both
subsets among the compared methods. Relative to LAPA-DINOv3, which has the
best average scores among the compared baselines, GeoLAM reduces the average
MSE from 0.60 to 0.29, a relative reduction of approximately 51.7\%, and
improves the average classification accuracy from 45.62\% to 71.22\%, a gain
of 25.60 percentage points. Classification accuracy reaches 74.19\% on human
videos and 68.25\% on robot videos. Although the latent-action model is
pretrained exclusively on action-free human videos, its representations
also support accurate prediction on the robot benchmarks. These results
indicate that GeoLAM encodes both control-relevant motion information and
high-level action semantics.

\begin{table}[!t]
    \centering
    \caption{Success rates (\%) on LIBERO~\cite{liu2023libero}.
    Best results among the compared methods are in bold.}
    \label{tab:libero}
    \small
    \setlength{\tabcolsep}{3pt}
    \renewcommand{\arraystretch}{1.10}
    \begin{tabular*}{\columnwidth}{@{\extracolsep{\fill}}l*{5}{c}@{}}
        \toprule
        Model & Spatial & Object & Goal & Long & Average \\
        \midrule
        $\pi_{0.5}$~\cite{intelligence2025pi05}
        & \textbf{98.8} & 98.2 & 98.0 & 92.4 & 96.9 \\
        OpenVLA-OFT~\cite{kim2025fine}
        & 97.6 & 98.4 & 97.9 & 94.5 & 97.1 \\
        Fast-WAM~\cite{yuan2026fastwam}
        & 98.2 & \textbf{100.0} & 97.0 & 95.2 & 97.6 \\
        Motus~\cite{bi2025motus}
        & 96.8 & 99.8 & 96.6 & 97.6 & 97.7 \\
        X-VLA~\cite{zheng2025xvla}
        & 98.2 & 98.6 & 97.8 & 97.6 & 98.1 \\
        LingBot-VA~\cite{li2026lingbotva}
        & 98.5 & 99.6 & 97.2 & \textbf{98.5} & \textbf{98.5} \\
        \midrule
        \textbf{Ours}
        & 98.5 & 99.8 & \textbf{98.2} & 97.8 & \textbf{98.5} \\
        \bottomrule
    \end{tabular*}
\end{table}

\begin{table}[!t]
    \centering
    \caption{Success rates (\%) on RoboTwin 2.0~\cite{chen2025robotwin2}.
    Rand.\ denotes randomized scenes.
    Best results among the compared methods are in bold.}
    \label{tab:robotwin}
    \small
    \setlength{\tabcolsep}{5pt}
    \renewcommand{\arraystretch}{1.10}
    \begin{tabular*}{\columnwidth}{@{\extracolsep{\fill}}l*{3}{c}@{}}
        \toprule
        Model & Clean & Rand. & Average \\
        \midrule
        X-VLA~\cite{zheng2025xvla} & 72.80 & 72.84 & 72.8 \\
        $\pi_{0.5}$~\cite{intelligence2025pi05} & 82.74 & 76.76 & 79.8 \\
        ABot-M0~\cite{yang2026abotm0} & 86.06 & 85.08 & 85.6 \\
        Motus~\cite{bi2025motus} & 88.66 & 87.02 & 87.8 \\
        Fast-WAM~\cite{yuan2026fastwam} & 91.88 & 91.78 & 91.8 \\
        LingBot-VA~\cite{li2026lingbotva} & 92.90 & 91.50 & 92.2 \\
        \midrule
        \textbf{Ours} & \textbf{93.00} & \textbf{92.20} & \textbf{92.6} \\
        \bottomrule
    \end{tabular*}
\end{table}

\subsection{Robotic Manipulation}

\begin{figure}[!t]
    \centering
    \includegraphics[width=0.9\linewidth]{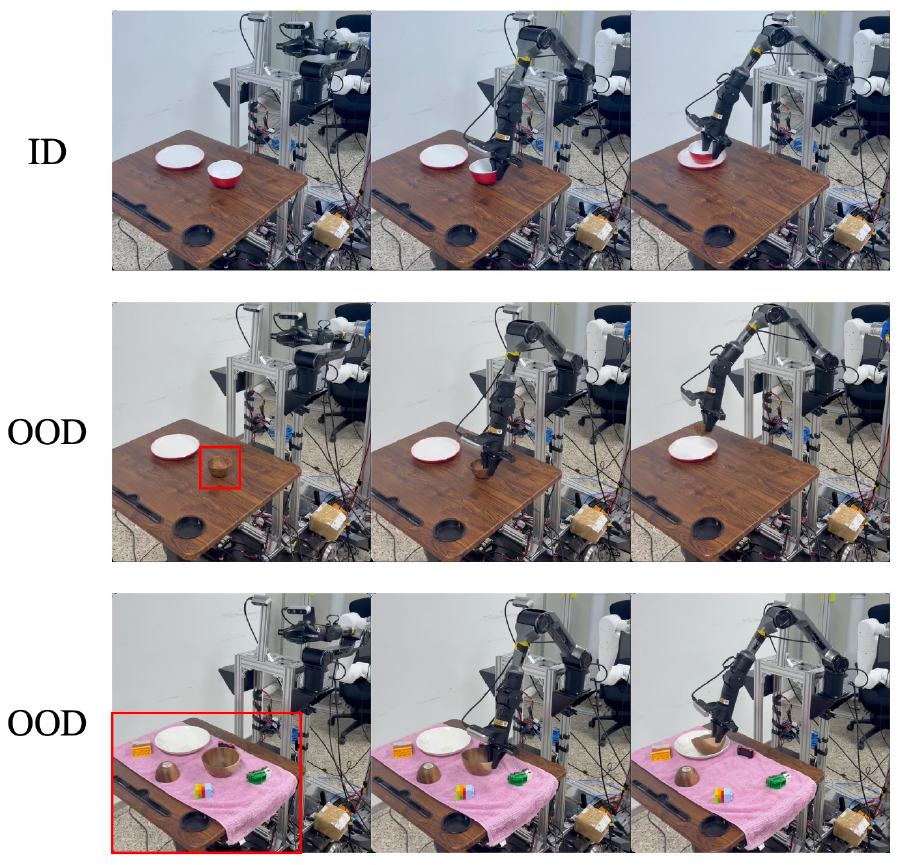}
    \caption{\textbf{Real-world bowl-placement task on Agilex Piper.}
    Representative executions of our policy, with time progressing from left
    to right. From top to bottom: the in-domain (ID) setting, an
    out-of-distribution (OOD) bowl, and an OOD scene with a different tabletop
    covering and additional distractors. Red boxes highlight the object and
    scene changes.}
    \label{fig:robot_1}
\end{figure}

\begin{figure}[!t]
    \centering
    \includegraphics[width=0.9\linewidth]{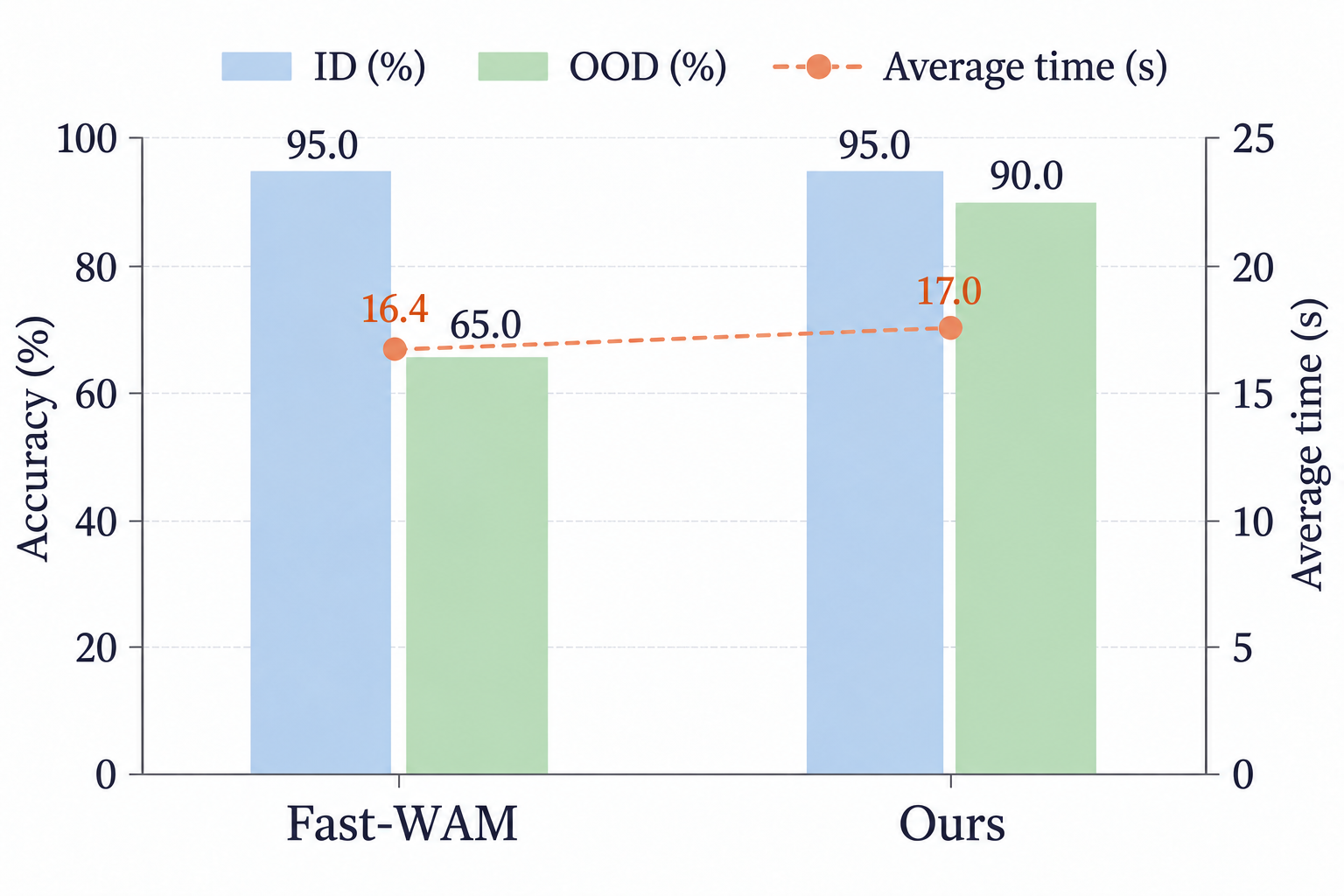}
    \caption{\textbf{Real-world success rates and execution time.}
    Bars show in-domain (ID) and out-of-distribution (OOD) task success rates
    (left axis), and the dashed line shows average execution time in seconds
    (right axis). Each success rate is based on 20 trials.
    Our policy matches Fast-WAM in the ID setting and improves OOD success
    by 25 percentage points.}
    \label{fig:robot_2}
\end{figure}

\subsubsection{LIBERO and RoboTwin Benchmarks}
\label{sec:simulation_benchmarks}

To assess the downstream utility of the learned latent actions, we evaluate
the resulting world action model on LIBERO~\cite{liu2023libero} and
RoboTwin 2.0~\cite{chen2025robotwin2}. We compare task success rates against
representative VLA and WAM baselines.

\noindent\textbf{LIBERO.}
LIBERO comprises four task suites covering spatial relations, object
variation, goal specification, and long-horizon manipulation.
Table~\ref{tab:libero} shows that our model achieves an average success
rate of 98.5\%, matching LingBot-VA~\cite{li2026lingbotva} and exceeding
X-VLA~\cite{zheng2025xvla} and Fast-WAM~\cite{yuan2026fastwam} by 0.4 and
0.9 percentage points, respectively. It achieves the best result on
Goal, with a success rate of 98.2\%, while remaining competitive on
Spatial, Object, and Long. The largest gain over Fast-WAM is on
Long, where success increases from 95.2\% to 97.8\%. These results support
the effectiveness of the GeoLAM-based policy on extended manipulation
sequences as well as shorter tasks.

\noindent\textbf{RoboTwin 2.0.}
RoboTwin 2.0 evaluates coordinated bimanual manipulation under clean and
randomized scene settings. As shown in Table~\ref{tab:robotwin}, our model
achieves success rates of 93.00\% and 92.20\% in these two settings,
respectively, yielding an average of 92.6\%. This exceeds
LingBot-VA~\cite{li2026lingbotva} and Fast-WAM by 0.4 and 0.8 percentage
points, respectively, and ranks first among the compared methods in both
settings. The success rate decreases by 0.8 percentage points under
randomization, indicating that the policy maintains high performance
across scene conditions. Together with the LIBERO results, this suggests
that the learned latent actions can support policy execution across
different manipulation tasks and embodiments.

\subsubsection{Real-World Experiments}
\label{sec:real_world_experiments}

We further evaluate the GeoLAM-based policy on an Agilex Piper robotic arm
performing a tabletop pick-and-place task: grasping a small bowl and placing
it on a plate. We use 100 real-world demonstration episodes for policy
training and evaluate each method in 20 in-domain (ID) trials and 20
out-of-distribution (OOD) trials. The ID setting matches the training
scenes, whereas the OOD evaluation covers changes to the manipulated object and
the scene background. Figure~\ref{fig:robot_1} shows
three representative execution sequences: the ID setup, a novel bowl, and
an OOD scene with a different tabletop covering and additional distractors.
The task objective remains unchanged across these conditions.

As summarized in Fig.~\ref{fig:robot_2}, both our policy and
Fast-WAM~\cite{yuan2026fastwam} achieve a 95.0\% success rate in the ID
setting. On the OOD evaluation, our policy reaches 90.0\%, compared with
65.0\% for Fast-WAM, an improvement of 25 percentage points. The success
rate therefore drops by only 5 percentage points from ID to OOD for our
policy, versus 30 percentage points for Fast-WAM. Average execution time
is 17.0\,s for our policy and 16.4\,s for Fast-WAM, a difference of
0.6\,s. These results show improved task success under the tested visual
distribution shifts with a similar execution duration, supporting the
utility of geometry-grounded latent actions for real-world manipulation.

\subsection{Ablation Study}
\label{sec:ablation}

\begin{table*}[!t]
    \centering
    \caption{Ablation of latent-action supervision on LARYBench~\cite{nie2026lary}.
    $F_0$ and $F_1$ denote the two shallow DA3 feature levels.
    Feature-reconstruction variants use only MSE on the indicated levels.
    Pixel reconstruction uses $\mathcal L_{\mathrm{pixel}}$ without the 4D teacher,
    while the full model additionally uses $\mathcal L_{\mathrm{geo}}$.
    Classification accuracy is in percent. Best results are in bold.}
    \label{tab:ablation}
    \small
    \setlength{\tabcolsep}{4pt}
    \renewcommand{\arraystretch}{1.10}
    \begin{tabular*}{\textwidth}{@{\extracolsep{\fill}}l*{5}{c}@{\hspace{10pt}}*{3}{c}@{}}
        \toprule
        \multirow{2}{*}{\textbf{Variant}}
        & \multicolumn{5}{c}{Regression MSE $\downarrow$}
        & \multicolumn{3}{c}{Classification Accuracy $\uparrow$} \\
        \cmidrule(lr){2-6} \cmidrule(lr){7-9}
        & CALVIN & VLABench & RoboCOIN & AgiBot & Avg.
        & Human & Robot & Avg. \\
        \midrule
        Feature reconstruction ($F_0$)
        & 0.50 & 0.23 & 0.79 & 0.79 & 0.57 & 67.48 & 62.73 & 65.10 \\
        Feature reconstruction ($F_1$)
        & 0.46 & 0.22 & 0.76 & 0.75 & 0.54 & 69.34 & 63.02 & 66.18 \\
        Feature reconstruction ($F_0+F_1$)
        & \textbf{0.22} & 0.44 & 0.66 & 0.64 & 0.49 & 68.93 & 64.24 & 66.58 \\
        Pixel reconstruction (w/o Geo Loss)
        & 0.41 & 0.13 & 0.66 & 0.65 & 0.46 & 70.88 & 66.64 & 68.76 \\
        \midrule
        \textbf{GeoLAM (full)}
        & 0.27 & \textbf{0.09} & \textbf{0.40} & \textbf{0.42}
        & \textbf{0.29} & \textbf{74.19} & \textbf{68.25} & \textbf{71.22} \\
        \bottomrule
    \end{tabular*}
\end{table*}

We examine the reconstruction objective and the 4D geometry teacher using
the LARYBench evaluation protocol in
Sec.~\ref{sec:latent_action_benchmark}.

\noindent\textbf{Reconstruction objective.}
The feature-reconstruction variants apply MSE only to the indicated
predicted DA3 features, without RGB reconstruction or geometry supervision.
The feature labels specify the supervision targets. The shallow-feature
state representation is retained. As shown in Table~\ref{tab:ablation},
$F_1$ provides a stronger single-level target than $F_0$, improving both
average MSE (0.54 vs.\ 0.57) and classification accuracy
(66.18\% vs.\ 65.10\%). Joint $F_0+F_1$ reconstruction obtains the best
average scores among the feature-reconstruction variants, with 0.49 MSE
and 66.58\% accuracy, although its regression gains are not uniform across
datasets. In particular, it achieves the lowest CALVIN error of 0.22,
compared with 0.27 for the full model. Pixel reconstruction through the
frozen geometric hierarchy and RGB decoder yields 0.46 average MSE and
68.76\% accuracy even without the teacher, indicating the value of the
decoded-image reconstruction constraint.

\noindent\textbf{4D geometric supervision.}
Adding the geometry teacher to pixel reconstruction reduces regression
error on all four robot datasets and improves classification accuracy on
both the human and robot subsets. Average MSE decreases from 0.46 to 0.29,
a relative reduction of approximately 37.0\%, while average classification
accuracy increases from 68.76\% to 71.22\%, a gain of 2.46 percentage
points. The full model therefore achieves the strongest aggregate results,
supporting the complementary role of explicit motion supervision beyond
the geometric prior already present in the reconstruction pathway.

\begin{figure}[!t]
    \centering
    \includegraphics[width=0.9\linewidth]{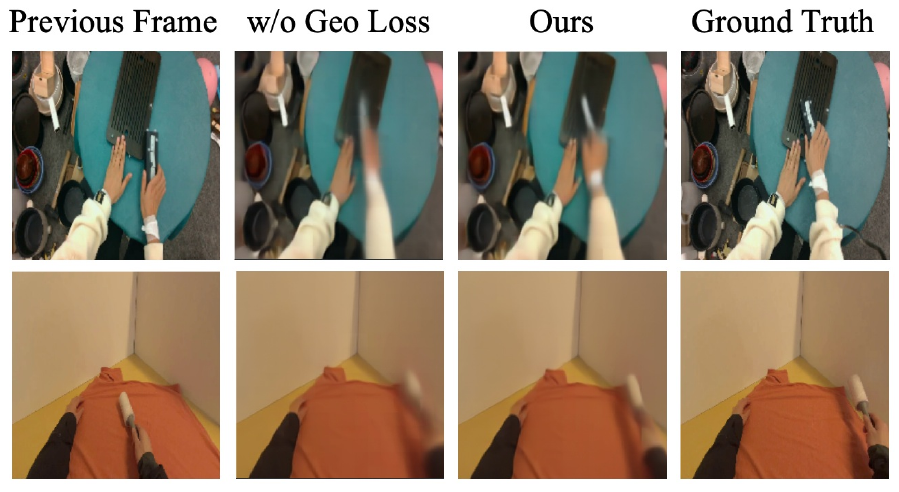}
    \caption{\textbf{Qualitative effect of 4D geometric supervision.}
    Each row shows a human-video transition. From left to right: the previous
    frame, reconstruction without D4RT supervision
    ($\mathcal L_{\mathrm{pixel}}$ only), full GeoLAM reconstruction, and the
    ground-truth future frame. Geometry supervision reduces smearing around
    moving hands and tools and better preserves their outlines.}
    \label{fig:ego}
\end{figure}

\noindent\textbf{Qualitative results.}
Figure~\ref{fig:ego} compares future-frame reconstructions for two human-video
transitions. Without geometry supervision, the moving hand and held tool
in the upper example have diffuse boundaries. The full model produces a
more distinct wrist boundary and tool outline. In the lower example, the
roller head is more compact and better separated from the cloth, while
the static background layout is similar in both variants. These
observations are consistent with supervision from displacement and
surface-orientation changes derived from D4RT~\cite{zhang2026d4rt}, whose
reliable motion cues are emphasized by the pooling scheme in
Sec.~\ref{sec:geometric_lal}. The qualitative improvement is concentrated
in dynamic regions, but fine texture and hand details remain imperfect.

\section{CONCLUSIONS}
In this paper, we presented GeoLAM, a framework for learning
geometry-grounded latent actions from action-free human videos and
transferring them to robot control. GeoLAM combines future-frame
reconstruction through a frozen geometric feature hierarchy with a
training-only 4D teacher that supervises 3D displacement, residual
image-plane motion, and surface-orientation changes. Reliability- and
motion-weighted spatial pooling encourages the continuous latent actions
to retain informative geometric motion without explicit hand-pose or
trajectory annotations. The frozen pretrained encoder supplies transition
targets for a world-action model that jointly denoises latent actions and
executable action chunks, while future-video prediction serves as an
auxiliary training objective. Deployment requires neither the geometry
teacher nor future-video generation. Results on LARYBench, LIBERO,
RoboTwin 2.0, and a real-world manipulation task support the utility of
the learned representation for motion and semantic prediction, as well
as downstream control.

Future work will focus on improving the reliability of teacher-derived
motion supervision under severe occlusion and complex camera motion.
We also plan to evaluate GeoLAM on more diverse real-world tasks and
robot embodiments, particularly long-horizon and contact-rich
manipulation, to better understand its transfer capabilities and
limitations.










\bibliographystyle{ieeetr}
\bibliography{references}

@inproceedings{xie2025universal,
  title     = {Universal Visuo-Tactile Video Understanding for Embodied Interaction},
  author    = {Xie, Yifan and Li, Mingyang and Li, Shoujie and Li, Xingting and Chen, Guangyu and Ma, Fei and Yu, Fei Richard and Ding, Wenbo},
  booktitle = {Advances in Neural Information Processing Systems},
  volume    = {38},
  pages     = {127864--127883},
  year      = {2025},
  publisher = {Curran Associates, Inc.},
  doi       = {10.52202/085713-4259},
  url       = {https://proceedings.neurips.cc/paper_files/paper/2025/file/b9a4d7b88a41652c63962ebcc21701b7-Paper-Conference.pdf}
}

@inproceedings{ye2025lapa,
  title     = {Latent Action Pretraining from Videos},
  author    = {Ye, Seonghyeon and Jang, Joel and Jeon, Byeongguk and Joo, Se June and Yang, Jianwei and Peng, Baolin and Mandlekar, Ajay and Tan, Reuben and Chao, Yu-Wei and Lin, Bill Yuchen and Liden, Lars and Lee, Kimin and Gao, Jianfeng and Zettlemoyer, Luke and Fox, Dieter and Seo, Minjoon},
  booktitle = {International Conference on Learning Representations},
  year      = {2025},
  url       = {https://proceedings.iclr.cc/paper_files/paper/2025/hash/45d74e190008c7bff2845ffc8e3facd3-Abstract-Conference.html}
}

@inproceedings{lin2026depthanything3,
  title     = {Depth Anything 3: Recovering the Visual Space from Any Views},
  author    = {Lin, Haotong and Chen, Sili and Liew, Jun Hao and Chen, Donny Y. and Li, Zhenyu and Zhao, Yang and Peng, Sida and Guo, Hengkai and Zhou, Xiaowei and Shi, Guang and Feng, Jiashi and Kang, Bingyi},
  booktitle = {International Conference on Learning Representations},
  year      = {2026},
  url       = {https://openreview.net/forum?id=yirunib8l8}
}

@article{jang2026gld,
  title         = {Repurposing Geometric Foundation Models for Multi-view Diffusion},
  author        = {Jang, Wooseok and Jeon, Seonghu and Han, Jisang and Choi, Jinhyeok and Kwon, Minkyung and Kim, Seungryong and Xie, Saining and Liu, Sainan},
  journal       = {arXiv preprint arXiv:2603.22275},
  year          = {2026},
  eprint        = {2603.22275},
  archiveprefix = {arXiv},
  primaryclass  = {cs.CV},
  doi           = {10.48550/arXiv.2603.22275},
  url           = {https://arxiv.org/abs/2603.22275}
}

@article{han2026gam,
  title         = {Geometric Action Model for Robot Policy Learning},
  author        = {Han, Jisang and Jeon, Seonghu and Jung, Jaewoo and Zurbr{\"u}gg, Ren{\'e} and An, Honggyu and Portela, Tifanny and Hutter, Marco and Pollefeys, Marc and Kim, Seungryong and Hong, Sunghwan},
  journal       = {arXiv preprint arXiv:2606.17046},
  year          = {2026},
  eprint        = {2606.17046},
  archiveprefix = {arXiv},
  primaryclass  = {cs.RO},
  doi           = {10.48550/arXiv.2606.17046},
  url           = {https://arxiv.org/abs/2606.17046}
}

@inproceedings{zhang2026d4rt,
  title     = {Efficiently Reconstructing Dynamic Scenes One {D4RT} at a Time},
  author    = {Zhang, Chuhan and Le Moing, Guillaume and Koppula, Skanda and Rocco, Ignacio and Momeni, Liliane and Xie, Junyu and Sun, Shuyang and Sukthankar, Rahul and Barral, Jo{\"e}lle K. and Hadsell, Raia and Ghahramani, Zoubin and Zisserman, Andrew and Zhang, Junlin and Sajjadi, Mehdi S. M.},
  booktitle = {Proceedings of the IEEE/CVF Conference on Computer Vision and Pattern Recognition},
  pages     = {7382--7392},
  year      = {2026},
  url       = {https://openaccess.thecvf.com/content/CVPR2026/html/Zhang_Efficiently_Reconstructing_Dynamic_Scenes_One_D4RT_at_a_Time_CVPR_2026_paper.html}
}

@inproceedings{bu2026laof,
  title     = {{LAOF}: Robust Latent Action Learning with Optical Flow Constraints},
  author    = {Bu, Xizhou and Lyu, Jiexi and Sun, Fulei and Yang, Ruichen and Ma, Zhiqiang and Li, Wei},
  booktitle = {Proceedings of the IEEE/CVF Conference on Computer Vision and Pattern Recognition},
  pages     = {27334--27344},
  year      = {2026},
  url       = {https://openaccess.thecvf.com/content/CVPR2026/html/Bu_LAOF_Robust_Latent_Action_Learning_with_Optical_Flow_Constraints_CVPR_2026_paper.html}
}

@article{yang2026lawm3d,
  title         = {{LAWM-3D}: Learning {3D-Aware} Latent Actions from Human Videos for Generalizable Robot World Models},
  author        = {Yang, Jiarui and Zhange, Jiale and Li, Jiawei and Guo, Hang and Huang, Wen and Wang, Jinpeng and Liu, Peidong and Xia, Shu-Tao},
  journal       = {arXiv preprint arXiv:2608.05706},
  year          = {2026},
  eprint        = {2608.05706},
  archiveprefix = {arXiv},
  primaryclass  = {cs.CV},
  doi           = {10.48550/arXiv.2608.05706},
  url           = {https://arxiv.org/abs/2608.05706}
}

@article{yuan2026fastwam,
  title         = {{Fast-WAM}: Do World Action Models Need Test-time Future Imagination?},
  author        = {Yuan, Tianyuan and Dong, Zibin and Liu, Yicheng and Zhao, Hang},
  journal       = {arXiv preprint arXiv:2603.16666},
  year          = {2026},
  eprint        = {2603.16666},
  archiveprefix = {arXiv},
  primaryclass  = {cs.CV},
  doi           = {10.48550/arXiv.2603.16666},
  url           = {https://arxiv.org/abs/2603.16666}
}

@article{shen2026ld4wam,
  title         = {{LD4WAM}: Learning Latent Dynamics from Human Videos for World Action Models},
  author        = {Shen, Zhenhao and Liang, Jiaqi and Lu, Jasper and Jiang, Feng and Wang, Yuran and Wei, Chuanbo and Liu, Jiayi and Yang, Jianchun and Yu, Qize and You, Jiadi and Hao, Ce and He, Guanqi and Xie, Chen and Wu, Ruihai},
  journal       = {arXiv preprint arXiv:2608.22403},
  year          = {2026},
  eprint        = {2608.22403},
  archiveprefix = {arXiv},
  primaryclass  = {cs.RO},
  doi           = {10.48550/arXiv.2608.22403},
  url           = {https://arxiv.org/abs/2608.22403}
}

@article{li2026lawa,
  title         = {Latent Action as Intention Enables Efficient Future Imagination for World Action Models},
  author        = {Li, Xiang and Zheng, Yupeng and Gu, Songen and Ma, Huailiang and Yu, Feng and Zheng, Yuhang and Nie, Xian and Yuan, Shanshuai and Zang, Yujie and Li, Weize and Tian, Shuai and Liu, Moyang and Zhang, Ya-Qin and Ding, Wenchao},
  journal       = {arXiv preprint arXiv:2608.24882},
  year          = {2026},
  eprint        = {2608.24882},
  archiveprefix = {arXiv},
  primaryclass  = {cs.RO},
  doi           = {10.48550/arXiv.2608.24882},
  url           = {https://arxiv.org/abs/2608.24882}
}

@inproceedings{peebles2023dit,
  title     = {Scalable Diffusion Models with Transformers},
  author    = {Peebles, William and Xie, Saining},
  booktitle = {Proceedings of the IEEE/CVF International Conference on Computer Vision},
  pages     = {4195--4205},
  year      = {2023},
  doi       = {10.1109/ICCV51070.2023.00387},
  url       = {https://openaccess.thecvf.com/content/ICCV2023/html/Peebles_Scalable_Diffusion_Models_with_Transformers_ICCV_2023_paper.html}
}

@inproceedings{lipman2023flow,
  title     = {Flow Matching for Generative Modeling},
  author    = {Lipman, Yaron and Chen, Ricky T. Q. and Ben-Hamu, Heli and Nickel, Maximilian and Le, Matthew},
  booktitle = {International Conference on Learning Representations},
  year      = {2023},
  url       = {https://openreview.net/forum?id=PqvMRDCJT9t}
}

@inproceedings{zhao2023act,
  title     = {Learning Fine-Grained Bimanual Manipulation with Low-Cost Hardware},
  author    = {Zhao, Tony Z. and Kumar, Vikash and Levine, Sergey and Finn, Chelsea},
  booktitle = {Proceedings of Robotics: Science and Systems},
  month     = {July},
  address   = {Daegu, Republic of Korea},
  year      = {2023},
  doi       = {10.15607/RSS.2023.XIX.016},
  url       = {https://www.roboticsproceedings.org/rss19/p016.html}
}

@article{wan2025,
  title         = {{Wan}: Open and Advanced Large-Scale Video Generative Models},
  author        = {{Wan Team} and Wang, Ang and Ai, Baole and Wen, Bin and others},
  journal       = {arXiv preprint arXiv:2503.20314},
  year          = {2025},
  eprint        = {2503.20314},
  archiveprefix = {arXiv},
  primaryclass  = {cs.CV},
  doi           = {10.48550/arXiv.2503.20314},
  url           = {https://arxiv.org/abs/2503.20314}
}

@misc{buildai2025egocentric10k,
  author       = {{Build AI}},
  title        = {Egocentric-10K},
  year         = {2025},
  howpublished = {Hugging Face Datasets},
  url          = {https://huggingface.co/datasets/builddotai/Egocentric-10K}
}

@article{punamiya2026egoverse,
  title         = {{EgoVerse}: An Egocentric Human Dataset for Robot Learning from Around the World},
  author        = {Punamiya, Ryan and Kareer, Simar and Liu, Zeyi and others},
  journal       = {arXiv preprint arXiv:2604.07607},
  year          = {2026},
  eprint        = {2604.07607},
  archiveprefix = {arXiv},
  primaryclass  = {cs.RO},
  doi           = {10.48550/arXiv.2604.07607},
  url           = {https://arxiv.org/abs/2604.07607}
}

@inproceedings{grauman2022ego4d,
  title     = {{Ego4D}: Around the World in 3,000 Hours of Egocentric Video},
  author    = {Grauman, Kristen and Westbury, Andrew and Byrne, Eugene and others},
  booktitle = {Proceedings of the IEEE/CVF Conference on Computer Vision and Pattern Recognition},
  pages     = {18995--19012},
  year      = {2022}
}

@article{li2026egolive,
  title         = {{EgoLive}: A Large-Scale Egocentric Dataset from Real-World Human Tasks},
  author        = {Li, Yihang and Wei, Xuelong and Luo, Jingzhou and others},
  journal       = {arXiv preprint arXiv:2604.23570},
  year          = {2026},
  eprint        = {2604.23570},
  archiveprefix = {arXiv},
  primaryclass  = {cs.RO},
  doi           = {10.48550/arXiv.2604.23570},
  url           = {https://arxiv.org/abs/2604.23570}
}

@inproceedings{hoque2026egodex,
  title     = {{EgoDex}: Learning Dexterous Manipulation from Large-Scale Egocentric Video},
  author    = {Hoque, Ryan and Huang, Peide and Yoon, David J. and Sivapurapu, Mouli and Zhang, Jian},
  booktitle = {International Conference on Learning Representations},
  year      = {2026},
  url       = {https://openreview.net/forum?id=FFxkFMU89E}
}

@inproceedings{goyal2017something,
  title     = {The ``Something Something'' Video Database for Learning and Evaluating Visual Common Sense},
  author    = {Goyal, Raghav and Ebrahimi Kahou, Samira and Michalski, Vincent and Materzynska, Joanna and Westphal, Susanne and Kim, Heuna and Haenel, Valentin and Fruend, Ingo and Yianilos, Peter and Mueller-Freitag, Moritz and Hoppe, Florian and Thurau, Christian and Bax, Ingo and Memisevic, Roland},
  booktitle = {Proceedings of the IEEE International Conference on Computer Vision},
  pages     = {5842--5850},
  year      = {2017}
}

@inproceedings{wang2023holoassist,
  title     = {{HoloAssist}: An Egocentric Human Interaction Dataset for Interactive {AI} Assistants in the Real World},
  author    = {Wang, Xin and Kwon, Taein and Rad, Mahdi and Pan, Bowen and Chakraborty, Ishani and Andrist, Sean and Bohus, Dan and Feniello, Ashley and Tekin, Bugra and Frujeri, Felipe Vieira and Joshi, Neel and Pollefeys, Marc},
  booktitle = {Proceedings of the IEEE/CVF International Conference on Computer Vision},
  pages     = {20270--20281},
  year      = {2023}
}

@article{damen2022epickitchens,
  title   = {Rescaling Egocentric Vision: Collection, Pipeline and Challenges for {EPIC-KITCHENS-100}},
  author  = {Damen, Dima and Doughty, Hazel and Farinella, Giovanni Maria and Furnari, Antonino and Kazakos, Evangelos and Ma, Jian and Moltisanti, Davide and Munro, Jonathan and Perrett, Toby and Price, Will and Wray, Michael},
  journal = {International Journal of Computer Vision},
  volume  = {130},
  pages   = {33--55},
  year    = {2022},
  doi     = {10.1007/s11263-021-01531-2}
}

@inproceedings{bu2025univla,
  title     = {Learning to Act Anywhere with Task-centric Latent Actions},
  author    = {Bu, Qingwen and Yang, Yanting and Cai, Jisong and Gao, Shenyuan and Ren, Guanghui and Yao, Maoqing and Luo, Ping and Li, Hongyang},
  booktitle = {Proceedings of Robotics: Science and Systems},
  year      = {2025},
  doi       = {10.15607/RSS.2025.XXI.014},
  url       = {https://www.roboticsproceedings.org/rss21/p014.html}
}

@inproceedings{chen2025villa,
  title     = {{villa-X}: Enhancing Latent Action Modeling in Vision-Language-Action Models},
  author    = {Chen, Xiaoyu and Wei, Hangxing and Zhang, Pushi and Zhang, Chuheng and Wang, Kaixin and Guo, Yanjiang and Yang, Rushuai and Wang, Yucen and Xiao, Xinquan and Zhao, Li and Chen, Jianyu and Bian, Jiang},
  booktitle = {International Conference on Learning Representations},
  year      = {2026},
  url       = {https://openreview.net/forum?id=y5CaJb17Fn}
}

@article{nie2026lary,
  title         = {{LARY}: A Latent Action Representation Yielding Benchmark for Generalizable Vision-to-Action Alignment},
  author        = {Nie, Dujun and Chen, Fengjiao and Lv, Qi and Kuang, Jun and Li, Xiaoyu and Cao, Xuezhi and Cai, Xunliang},
  journal       = {arXiv preprint arXiv:2604.11689},
  year          = {2026},
  eprint        = {2604.11689},
  archiveprefix = {arXiv},
  primaryclass  = {cs.CV},
  doi           = {10.48550/arXiv.2604.11689},
  url           = {https://arxiv.org/abs/2604.11689}
}

@article{liu2023libero,
  title   = {{LIBERO}: Benchmarking Knowledge Transfer for Lifelong Robot Learning},
  author  = {Liu, Bo and Zhu, Yifeng and Gao, Chongkai and Feng, Yihao and Liu, Qiang and Zhu, Yuke and Stone, Peter},
  journal = {arXiv preprint arXiv:2306.03310},
  year    = {2023},
  url     = {https://arxiv.org/abs/2306.03310}
}

@article{chen2025robotwin2,
  title   = {{RoboTwin 2.0}: A Scalable Data Generator and Benchmark with Strong Domain Randomization for Robust Bimanual Robotic Manipulation},
  author  = {Chen, Tianxing and Chen, Zanxin and Chen, Baijun and Cai, Zijian and Liu, Yibin and Li, Zixuan and Liang, Qiwei and Lin, Xianliang and Ge, Yiheng and Gu, Zhenyu and Deng, Weiliang and Guo, Yubin and Nian, Tian and Xie, Xuanbing and Chen, Qiangyu and Su, Kailun and Xu, Tianling and Liu, Guodong and Hu, Mengkang and Gao, Huan-ang and Wang, Kaixuan and Liang, Zhixuan and Qin, Yusen and Yang, Xiaokang and Luo, Ping and Mu, Yao},
  journal = {arXiv preprint arXiv:2506.18088},
  year    = {2025},
  url     = {https://arxiv.org/abs/2506.18088}
}

@article{intelligence2025pi05,
  title   = {{\(\pi_{0.5}\)}: A Vision-Language-Action Model with Open-World Generalization},
  author  = {{Physical Intelligence} and Black, Kevin and Brown, Noah and others},
  journal = {arXiv preprint arXiv:2504.16054},
  year    = {2025},
  url     = {https://arxiv.org/abs/2504.16054}
}

@article{kim2025fine,
  title   = {Fine-Tuning Vision-Language-Action Models: Optimizing Speed and Success},
  author  = {Kim, Moo Jin and Finn, Chelsea and Liang, Percy},
  journal = {arXiv preprint arXiv:2502.19645},
  year    = {2025},
  url     = {https://arxiv.org/abs/2502.19645}
}

@article{bi2025motus,
  title   = {{Motus}: A Unified Latent Action World Model},
  author  = {Bi, Hongzhe and Tan, Hengkai and Xie, Shenghao and Wang, Zeyuan and Huang, Shuhe and Liu, Haitian and Zhao, Ruowen and Feng, Yao and Xiang, Chendong and Rong, Yinze and Zhao, Hongyan and Liu, Hanyu and Su, Zhizhong and Ma, Lei and Su, Hang and Zhu, Jun},
  journal = {arXiv preprint arXiv:2512.13030},
  year    = {2025},
  url     = {https://arxiv.org/abs/2512.13030}
}

@inproceedings{zheng2025xvla,
 author = {Zheng, Jinliang and Li, Jianxiong and Wang, Zhihao and Liu, Dongxiu and Kang, Xirui and Feng, Yuchun and Zheng, Yinan and Zou, Jiayin and Chen, Yilun and Zeng, Jia and Wang, Tai and Zhang, Ya-Qin and Liu, Jingjing and Zhan, Xianyuan},
 booktitle = {International Conference on Learning Representations},
 editor = {C. Vondrick and B. Hariharan and C. Raffel and L. Pinto and D. Yang and A. Faust},
 pages = {60580--60606},
 title = {X-VLA: Soft-Prompted Transformer as Scalable Cross-Embodiment Vision-Language-Action Model},
 url = {https://proceedings.iclr.cc/paper_files/paper/2026/file/630d293833e09e1ecd892a898a20b074-Paper-Conference.pdf},
 volume = {2026},
 year = {2026}
}

@article{yang2026abotm0,
  title   = {{ABot-M0}: {VLA} Foundation Model for Robotic Manipulation with Action Manifold Learning},
  author  = {Yang, Yandan and Zeng, Shuang and Lin, Tong and Chang, Xinyuan and Qi, Dekang and Xiao, Junjin and Liu, Haoyun and Chen, Ronghan and Chen, Yuzhi and Huo, Dongjie and Xiong, Feng and Wei, Xing and Ma, Zhiheng and Xu, Mu},
  journal = {arXiv preprint arXiv:2602.11236},
  year    = {2026},
  url     = {https://arxiv.org/abs/2602.11236}
}

@INPROCEEDINGS{li2026lingbotva, 
    AUTHOR    = {Lin Li AND Qihang Zhang AND Yiming Luo AND Shuai Yang AND Ruilin Wang AND Luyao Zhang AND Mingrui Yu AND Zelin Gao AND Nan Xue AND Boyu Zhou AND Xing Zhu AND Mingyu Ding AND Yujun Shen AND Yinghao Xu}, 
    TITLE     = {{Causal World Modeling for Robot Control}}, 
    BOOKTITLE = {Proceedings of Robotics: Science and Systems}, 
    YEAR      = {2026}, 
    ADDRESS   = {Sydney, Australia}, 
    MONTH     = {July}, 
    DOI       = {10.15607/RSS.2026.XXII.016} 
}

@inproceedings{zhang2025hawor,
  title     = {{HaWoR}: World-Space Hand Motion Reconstruction from Egocentric Videos},
  author    = {Zhang, Jinglei and Deng, Jiankang and Ma, Chao and Potamias, Rolandos Alexandros},
  booktitle = {Proceedings of the IEEE/CVF Conference on Computer Vision and Pattern Recognition},
  pages     = {1805--1815},
  month     = jun,
  year      = {2025},
  url       = {https://openaccess.thecvf.com/content/CVPR2025/html/Zhang_HaWoR_World-Space_Hand_Motion_Reconstruction_from_Egocentric_Videos_CVPR_2025_paper.html}
}

@article{team2026xiaomi,
  title         = {{Xiaomi-Robotics-1}: Scaling Vision-Language-Action Models with over {100K} Hours of Real-World Trajectories},
  author        = {{Xiaomi Robotics Team} and Guo, Jun and Jin, Piaopiao and Li, Jason and Li, Peiyan and Li, Yingyan and Liu, Futeng and Peng, Wanli and Qin, Optimus and Su, Yifei and Sun, Nan and Sun, Qiao and Suo, Runze and Wang, Heyun and Wang, Yunhong and Wu, Rujie and Xia, Caoyu and Zhang, Lina and Zhao, Jack and Chen, Guoliang and Chen, Wenlong and He, Xinze and Li, Bin and Li, Qing and Li, Zhuorong and Qu, Heng and Song, Wenxuan and Xiang, Diyun and Xie, Yifan and Xu, Peiran and Ye, Hangjun and Ye, Wen and Zhao, Han and Zhou, Quanyun},
  journal       = {arXiv preprint arXiv:2607.15330},
  year          = {2026},
  month         = jul,
  eprint        = {2607.15330},
  archiveprefix = {arXiv},
  primaryclass  = {cs.RO},
  doi           = {10.48550/arXiv.2607.15330},
  url           = {https://arxiv.org/abs/2607.15330}
}

@article{xie2026humanintention,
  title         = {Learning Human-Intention Priors from Large-Scale Human Demonstrations for Robotic Manipulation},
  author        = {Xie, Yifan and Wang, YuAn and Chen, Guangyu and Liu, Jinkun and Sun, Yu and Ding, Wenbo},
  journal       = {arXiv preprint arXiv:2604.24681},
  year          = {2026},
  eprint        = {2604.24681},
  archiveprefix = {arXiv},
  primaryclass  = {cs.RO},
  doi           = {10.48550/arXiv.2604.24681},
  url           = {https://arxiv.org/abs/2604.24681}
}

@article{liu2026oawam,
  title         = {{OA-WAM}: Object-Addressable World Action Model for Robust Robot Manipulation},
  author        = {Liu, Yushan and Sun, Peibo and Li, Shoujie and Xie, Yifan and Zhang, Lingfeng and Chao, Xintao and Dong, Shiyuan and Chen, Fang and Zhang, Xiao-Ping and Ding, Wenbo},
  journal       = {arXiv preprint arXiv:2605.06481},
  year          = {2026},
  month         = may,
  eprint        = {2605.06481},
  archiveprefix = {arXiv},
  primaryclass  = {cs.RO},
  doi           = {10.48550/arXiv.2605.06481},
  url           = {https://arxiv.org/abs/2605.06481}
}

@article{hu2024jepa3d,
  title         = {{3D-JEPA}: A Joint Embedding Predictive Architecture for {3D} Self-Supervised Representation Learning},
  author        = {Hu, Naiwen and Cheng, Haozhe and Xie, Yifan and Li, Shiqi and Zhu, Jihua},
  journal       = {arXiv preprint arXiv:2409.15803},
  year          = {2024},
  eprint        = {2409.15803},
  archiveprefix = {arXiv},
  primaryclass  = {cs.CV},
  doi           = {10.48550/arXiv.2409.15803},
  url           = {https://arxiv.org/abs/2409.15803}
}

@article{li2026mastermicro,
  title         = {Master Micro Residual Correction with Adaptive Tactile Fusion and Force-Mixed Control for Contact-Rich Manipulation},
  author        = {Li, Xingting and Xie, Yifan and Liu, Han and Hou, Wei and Chen, Guangyu and Li, Shoujie and Ding, Wenbo},
  journal       = {arXiv preprint arXiv:2603.15152},
  year          = {2026},
  eprint        = {2603.15152},
  archiveprefix = {arXiv},
  primaryclass  = {cs.RO},
  doi           = {10.48550/arXiv.2603.15152},
  url           = {https://arxiv.org/abs/2603.15152}
}

\end{document}